\documentclass[letterpaper, 10 pt, conference]{ieeeconf}  

\usepackage{times}

\IEEEoverridecommandlockouts                              

\usepackage[T1]{fontenc}
\usepackage[utf8]{inputenc}

\usepackage{newtxtext,newtxmath}

\usepackage{times}
\usepackage{mathtools}

\usepackage{multicol}
\usepackage[bookmarks=true,backref=section,hidelinks]{hyperref} 
\usepackage{cleveref}
\usepackage{csquotes} 
\usepackage{graphicx}
\usepackage{silence}
\usepackage[caption=false,font=footnotesize]{subfig}
\usepackage{mathtools}
\usepackage{wrapfig}
\usepackage[table]{xcolor} 
\usepackage[ruled,vlined]{algorithm2e}
\usepackage[
  expansion=true,
  protrusion=false, 
  final
]{microtype}
\usepackage{booktabs}
\usepackage{flushend}

\definecolor{lightcyan}{rgb}{0.9, 0.98, 0.98} %

\begin{document}

\title{\LARGE \bf Stability-Guided Exploration for Diverse Motion Generation}

\author{Eckart Cobo-Briesewitz$^{*1}$, Tilman Burghoff$^{1}$, Denis Shcherba$^{1}$, Armand Jordana$^{2}$ and Marc Toussaint$^{1,3}$
\thanks{$^{*}$Corresponding Author: cobo-briesewitz@campus.tu-berlin.de}
\thanks{$^{1}$Technische Universität Berlin}
\thanks{$^{2}$LAAS-CNRS}
\thanks{$^{3}$Robotics Institute Germany (RIG)}}



\maketitle

\thispagestyle{empty}
\pagestyle{empty}

\begin{abstract}
Scaling up datasets is highly effective in improving the performance of deep learning models, including in the field of robot learning. However, data collection still proves to be a bottleneck. Approaches relying on collecting human demonstrations are labor-intensive and inherently limited: they tend to be narrow, task-specific, and fail to adequately explore the full space of feasible states. Synthetic data generation could remedy this, but current techniques mostly rely on local trajectory optimization and fail to find diverse solutions. In this work, we propose a novel method capable of finding diverse long-horizon manipulations through black-box simulation. We achieve this by combining an RRT-style search with sampling-based MPC, together with a novel sampling scheme that guides the exploration toward stable configurations. Specifically, we sample from a manifold of stable states while growing a search tree directly through simulation, without restricting the planner to purely stable motions. We demonstrate the method’s ability to discover diverse manipulation strategies, including pushing, grasping, pivoting, throwing, and tool use, across different robot morphologies, without task-specific guidance.
\end{abstract}

\IEEEpeerreviewmaketitle

\section{Introduction}
\subsection{Motivation}

Recent advances in deep learning have demonstrated impressive capabilities for controlling robots~\cite{zitkovich2023RT2, driess2023palme, 2025pi05}. These developments emphasize the need for large-scale, diverse robotic datasets to enable further progress.
However, compared to domains like vision and language, datasets for robotics are not as readily available. This has led to research efforts directed towards obtaining more data. For this purpose, several different approaches have been explored. One of the most prominent in the field of behavioral cloning is teleoperation~\cite{fang2025airexo, ze2025twist, zhao2023aloha}, where a human operator controls a robot to solve a specific task. While this can lead to high quality real world data, it requires extensive work from human experts, which is both expensive and time consuming. 
For reference, current large scale robotic datasets contain around a few million demonstrations~\cite{openx}, whereas VLM datasets comprise billions of image-text pairs~\cite{schuhmann2022laion}.
An alternative approach is to extract data 
from human video demonstrations~\cite{liu2025immimic, xu2025dexumi}, 
by using computer vision to estimate the positions of humans directly from online videos~\cite{McCarthy2025}. The motivation for this type of method is the large amount of human videos found on the internet.
However, these techniques do not produce the necessary low-level information needed for training robots.
Furthermore, they only focus on human-like approaches, neglecting the diversity of robot morphologies.
%

Overall, human generated data is inherently limited, as it may not be optimal and can fail to capture the full range of solutions accessible to robots. In contrast, algorithmically generated synthetic data can provide a broader set of valid solutions.
Consequently, in this work, we are interested in the use of simulators to directly create high-quality and diverse data on tasks involving complex robot-object interactions. 

Sampling-based Model Predictive Control (MPC) has recently shown promising results in generating dynamic motions by directly interacting with the simulator, both in locomotion~\cite{xue2025full} and manipulation~\cite{li2025drop}. However, these techniques mostly rely on local exploration in the control space~\cite{jordana2025introductionzeroorderoptimizationtechniques}, and are therefore prone to local minima.
In contrast, sampling-based planners~\cite{lavalle1998rapidlyexploring, lavalle1999randomized} have a long history of relying on state space sampling to explore globally.
We aim to combine both approaches to generate diverse, dynamic and contact-rich manipulations.

\begin{figure}[t]
    \centering
    \includegraphics[width=\linewidth]{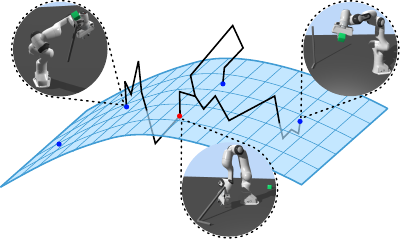}
    \caption{Schematic of the paths found by our method. For each start node (red), we grow a tree \textit{guided} by the manifold of stable states, but not constrained by it.}
    \label{fig:visualAbstract}
\end{figure}

\begin{figure*}[t]
    \centering
    \subfloat[SpheresRamp]{\includegraphics[width=0.22\textwidth]{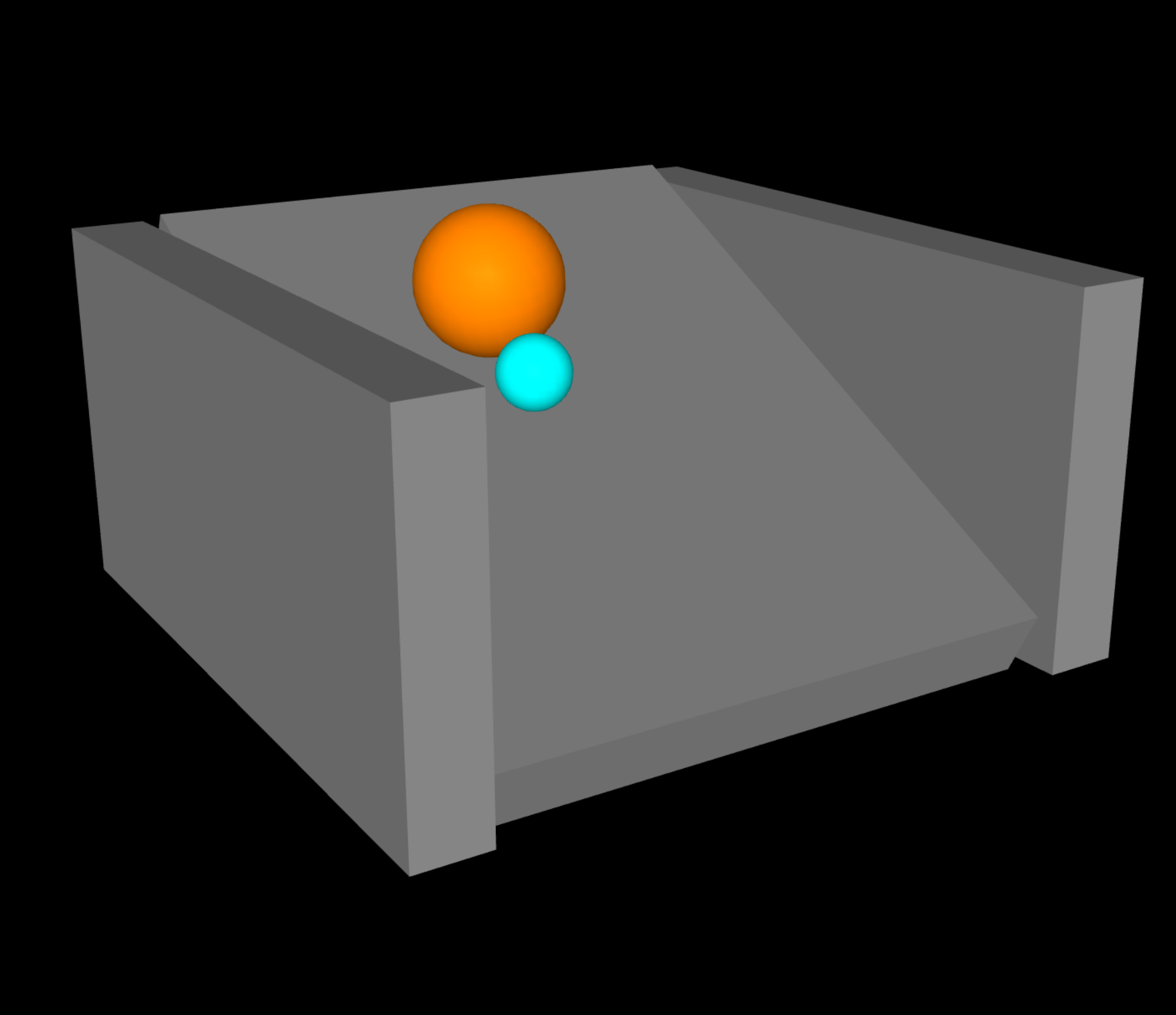}} \hfill
    \subfloat[SpheresCube]{\includegraphics[width=0.22\textwidth]{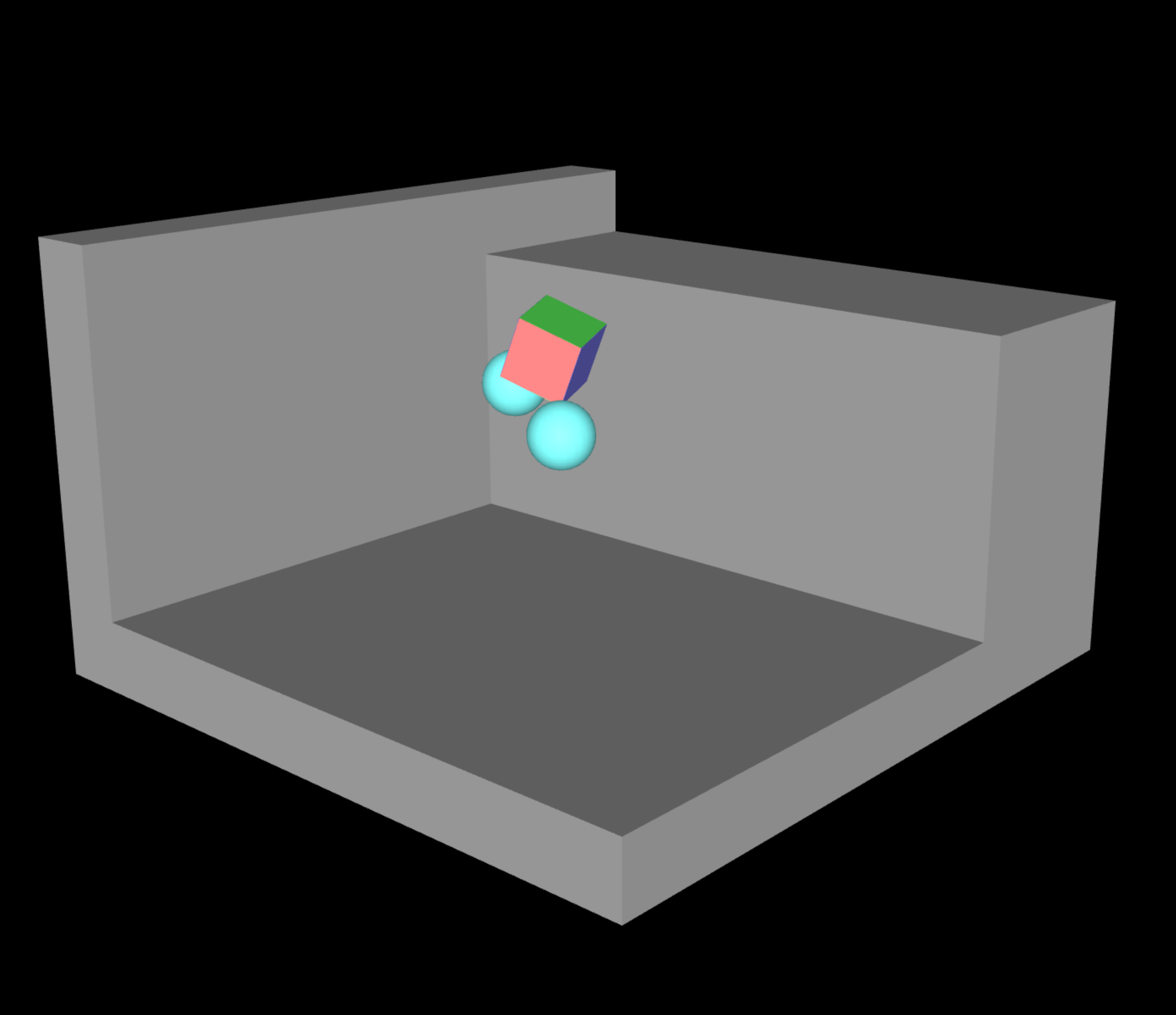}} \hfill
    \subfloat[PandaHook]{\includegraphics[width=0.22\textwidth]{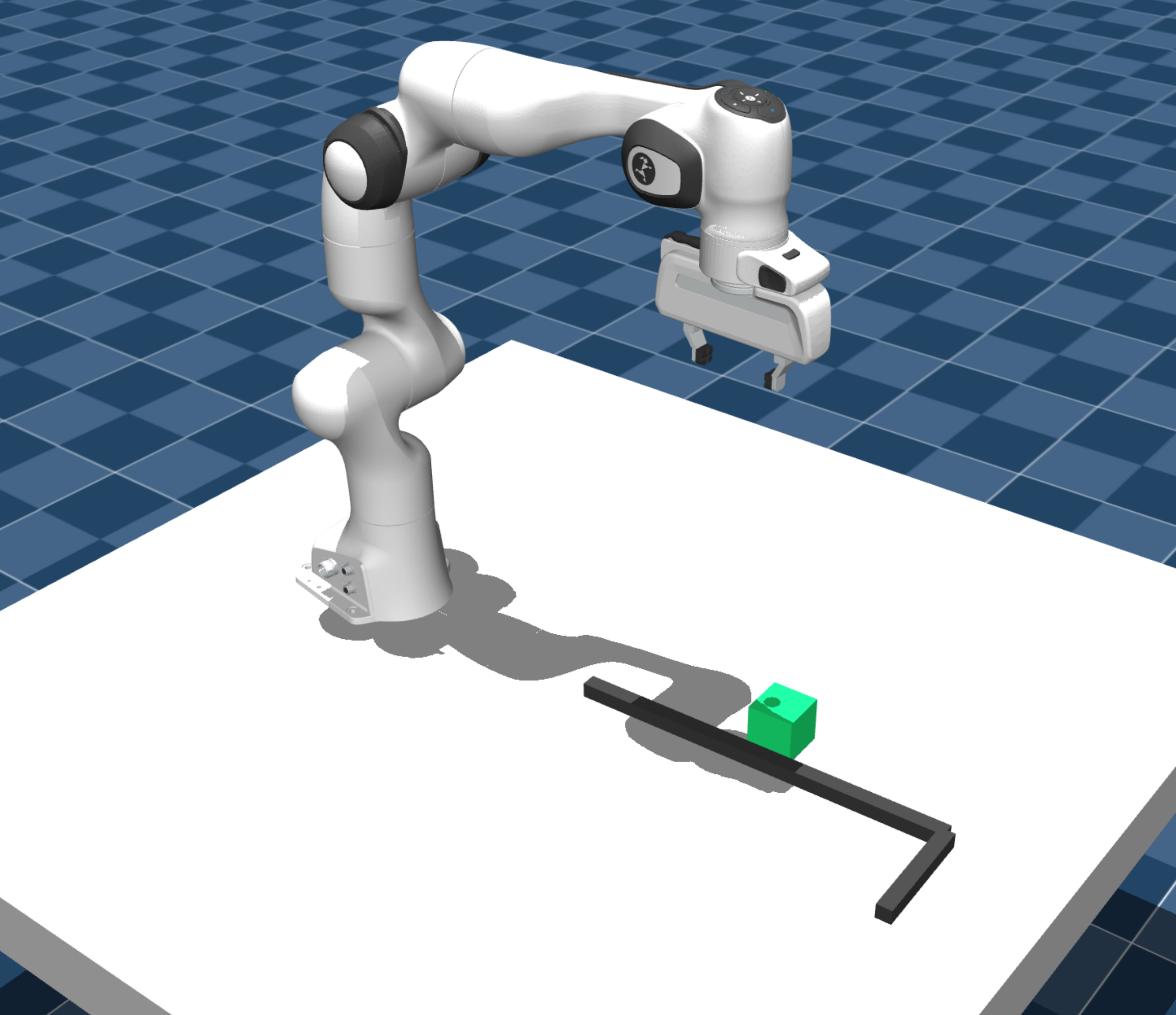}} \hfill
    \subfloat[PandasCube]{\includegraphics[width=0.22\textwidth]{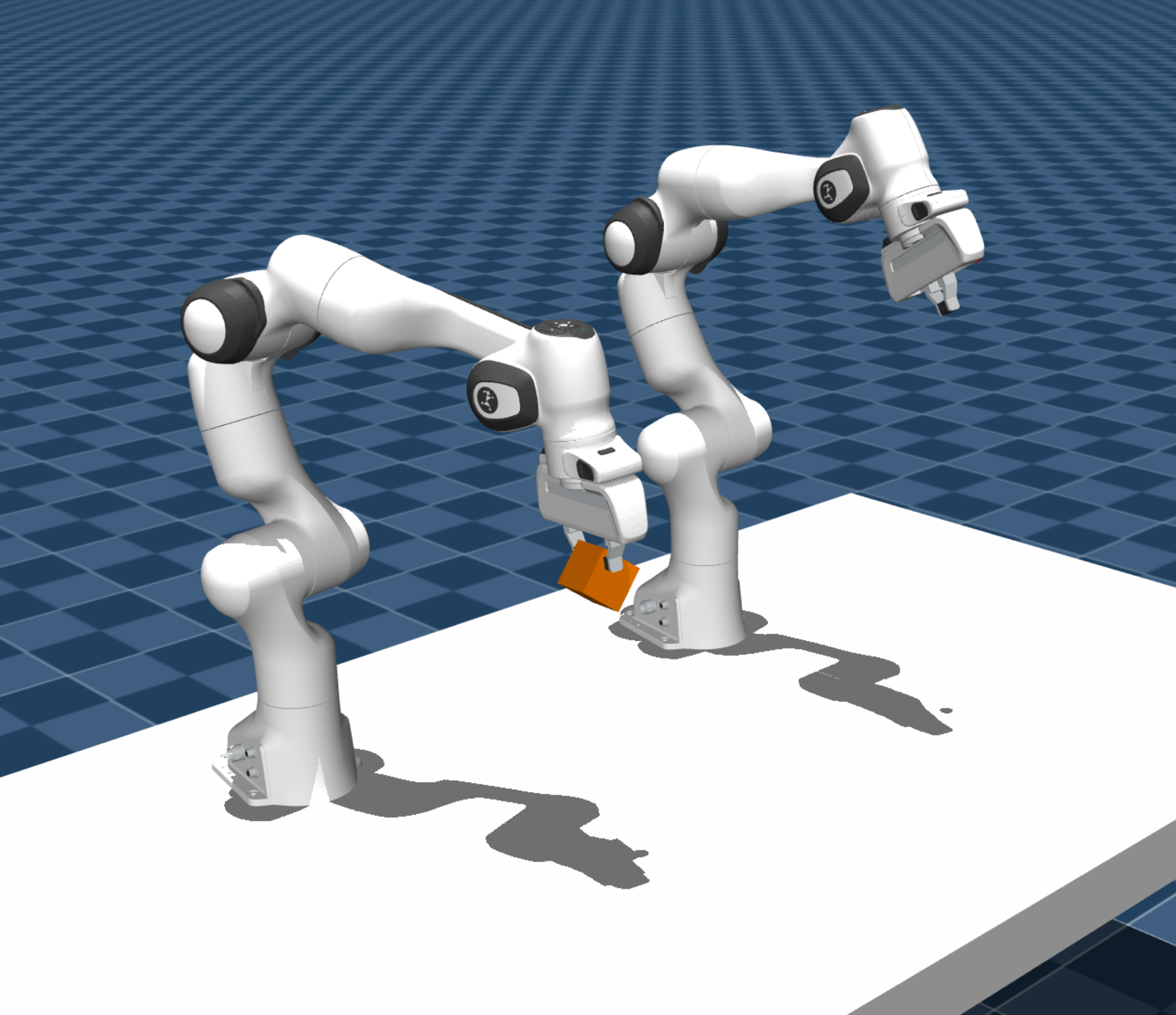}}
    \caption{The four environments used in our experiments. For the first two scenes the blue balls represent robots containing a translational joint in 3D space. The complexity of (a) lies in the high likelihood of landing in local minima, as once the object (orange) falls off the ramp it is impossible to recover. Environment (b) tests how well the algorithm can find diverse paths using two possible robot contact points as well as having a state where the rotation of the object is highly relevant. In (c) the method is tested for the ability of finding paths containing tool use. And environment (d) test the method on a high dimensional space where two robots can cooperate with each other to manipulate an object.}
    \label{fig:environments}
\end{figure*}

\subsection{Related Work}
\paragraph{Sampling-based MPC}

MPC defines a controller as the solution to an optimal control problem. While MPC in robotics has historically relied on gradient-based optimization techniques~\cite{mastalli2020crocoddyl, verschueren2022acados}, recent advances in parallel computing have enabled the successful deployment of gradient-free techniques in MPC on robots~\cite{xue2025full, li2025drop}. In practice, most techniques currently used in robotics rely on some form of iterative procedure where samples following a Gaussian distribution are used to evaluate the function around the current guess in order to find a descent direction~\cite{jordana2025introductionzeroorderoptimizationtechniques}. For instance, Predictive Sampling simply picks the best sample to be the new guess around which the next Gaussian samples are drawn~\cite{howell2022predictive}. MPPI uses a softmax weighting of the samples to compute an update direction~\cite{williams2016aggressive}. CMA-ES relies on a weighted average of the $k$-best samples (and also iteratively updates the covariance matrix)~\cite{hansen2001completely}.

The main strength of these techniques is that they can find interesting robot motions simply by interacting with the simulator. 
However, these approaches are inherently local. In many MPC and trajectory optimization methods, sampling is performed in the control space, often using single-shooting formulations, as handling constraints directly in gradient-free optimization is difficult. In contrast, graph-based motion planning methods sample in the state space and can offer theoretical guarantees such as probabilistic completeness.


\paragraph{Sampling Based Motion Planning}

Sampling-based motion planners solve motion planning problems using probabilistic methods. They are commonly divided into two categories: \textit{single-query motion planning}, with the most prominent approach being Rapidly-exploring Random Trees (RRTs)~\cite{lavalle1998rapidlyexploring}, and \textit{multi-query motion planning}, which is typically addressed using probabilistic roadmaps (PRMs)~\cite{kavraki1996probabilistic}. 
RRTs incrementally build a tree by sampling random states and extending the nearest node in the tree toward each sample. In contrast, PRMs build a graph by sampling points in the configuration space and attempting to connect nearby samples. This roadmap can then be reused to efficiently answer motion planning queries online.
These algorithms are probabilistically complete under mild assumptions~\cite{LaValle2006},
meaning that they find a solution, if it exists, as the number of samples tends to infinity.

These methods have proven effective in many domains and have led to various extensions being developed \cite{kuffner2000rrtconnect,karaman2011samplingbased,gammell2014informed}. 
In the context of our work, an important extension is manifold RRTs~\cite{berenson2009manipulation, suh2011tangent}. When constraints imposed on the system define a manifold of lower dimension than the ambient space, the probability of finding feasible points through uniform sampling in the ambient space approaches zero. Manifold RRTs address this issue by projecting sampled points onto the manifold and modifying the extend step to ensure that the tree grows along it.

In our work, we use the manifold of stable configurations to guide the search. Similarly to manifold RRT, we consider points projected onto a manifold as targets for extension; however, we do not restrict the paths to remain within the manifold. In that sense, our method is guided by a manifold of stable states while still being able to explore unstable states outside of it. This enables us to find diverse non-prehensile manipulations.

\paragraph{Kinodynamic planning} Kinodynamic planning aims to solve problems involving both kinematics and dynamic constraints. 
The most direct approach is to use trajectory optimization~\cite{mayne1966second}; however, similarly to sampling-based MPC, it is subject to local minima and therefore cannot solve tasks involving long horizons. Another approach is to use Task and Motion Planning (TAMP)~\cite{garrett2021integrated}; however, this implies a tedious manual definition of predicates.
Alternatively, one could use sampling-based planners. For instance, RRTs can naturally be used for kino-dynamic planning by modifying the extend step to select a control input, typically by sampling multiple candidates and keeping the one that yields a next state closest to the target~\cite{lavalle1999randomized}. 
Some works \cite{choudhury2016regionally, ortiz2024idb} combined trajectory optimization and RRTs; however, they rely on gradient based optimization, which limits their application to problems involving complex contact interactions.
In the context of manipulation, \cite{Barry2013, king2016rearrangement} combined RRT and motion primitives to perform non-prehensile manipulation. In \cite{cheng2022contact} contacts are used for guiding motion planning, but is restricted to quasidynamic manipulation.

Conceptually similar to our work, \cite{Haustein} uses stability to guide a kino-dynamic RRT; however, 
they only consider non-prehensile manipulations on a $2$-dimensional plane.
Furthermore, the authors enforce that the system ends up in a stable state after each action, while our approach allows the robots to execute multiple actions without intermittent stability.

\begin{figure*}[t]
    \vspace{0.5cm}
    \centering
    \includegraphics[width=\linewidth]{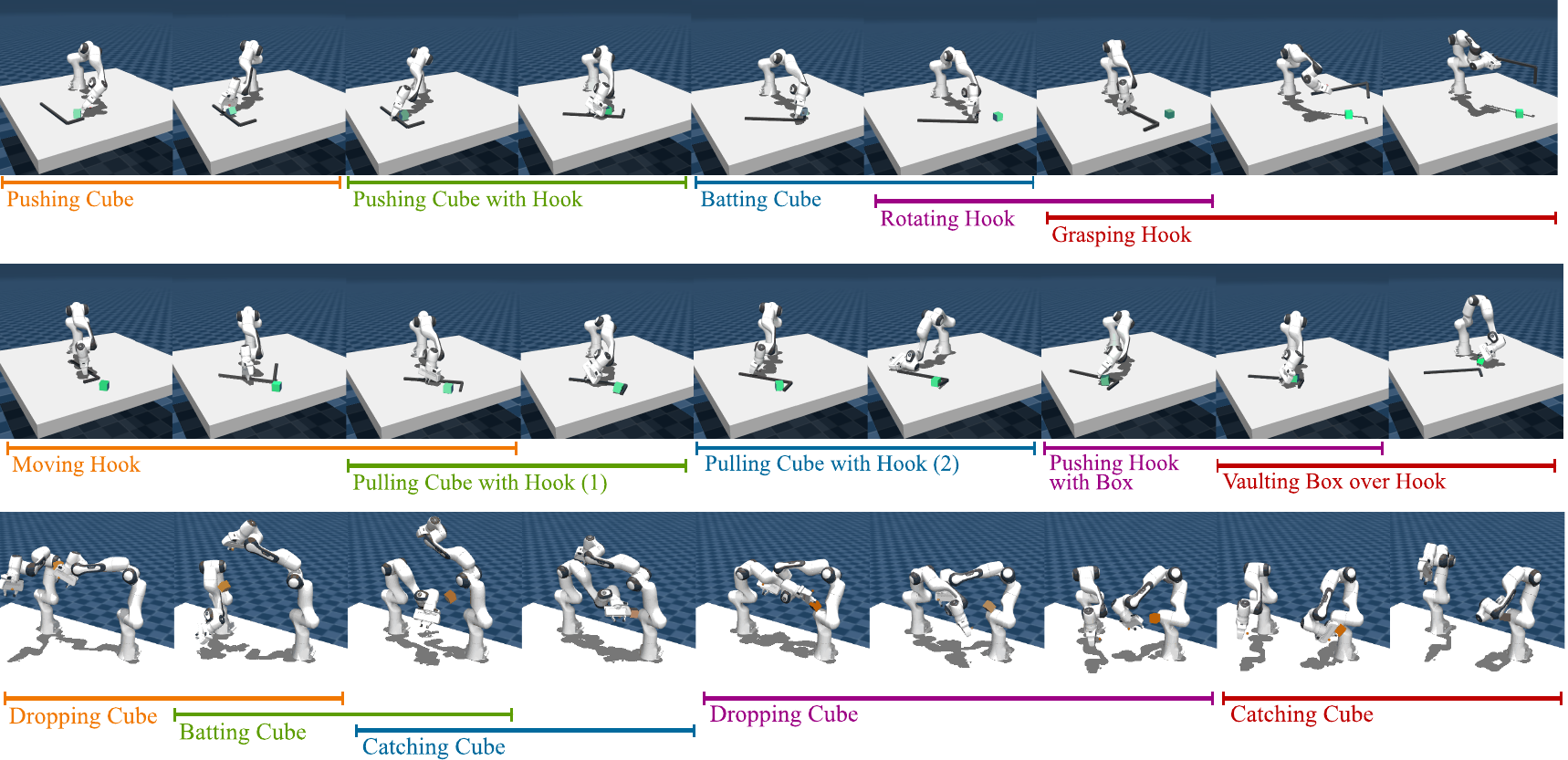}
    \caption{Examples of trajectories generated by \textit{StaGE}. The first trajectory shows diverse manipulations in the \textit{PandaHook} environment, while the second one demonstrates tool use (pulling the cube with the hook). The third trajectory shows a transfer of the cube from the left panda to the right, by throwing and catching the cube multiple times.}
    \label{fig:trajectories}
\end{figure*}
\subsection{Contributions}

Our contributions are as follows:
\begin{enumerate}
    \item[(i)] We introduce \textit{StaGE}, a novel
    algorithm to find complex and diverse long-horizon manipulations without motion priors. 
    We achieve this by using our novel sampling scheme, \textit{Stability-Guidance}, to guide a kino-dynamic RRT that directly interacts with black-box simulation. In addition, we provide a set of extensions to kino-dynamic RRT to encourage the generation of diverse behaviors.
    \item[(ii)] Evaluations in multiple challenging environments with different robot morphologies qualitatively demonstrate the nature of the manipulations found by our approach.
\end{enumerate}

Importantly, the proposed method is task-agnostic. All the obtained motions naturally emerge and do not rely on manually tuned cost functions.
To the best of our knowledge, this is the first generic method applying RRT with black-box simulation to non-prehensile manipulation without the use of hand-crafted motion primitives or analytical constraints.

\section{Method}

The aim of our method is to find diverse, contact-rich manipulations in any scene, independent of any task. To that end, we consider a hierarchy of subspaces $\mathcal{C}_{stable}\subset\mathcal{C}_{feasible}$, where $\mathcal{C}_{feasible}$ describes the space of all reachable states in a given black-box simulation, and $\mathcal{C}_{stable}$ additionally imposes that the state is stable, i.e., all objects are in equilibrium.  
%
Our method is divided into two stages. First, we sample states from $\mathcal{C}_{stable}$ using a diverse constraint solver. In the second stage, we use an RRT-style planner to find diverse paths between states. 
Importantly, the sampled random states are only used to guide the search; that is to say, the RRT-style planner is free to evolve through non-stable regions to enable dynamic manipulation. We provide a visual intuition for this approach in Fig.~\ref{fig:visualAbstract}.
The pseudo-code for \textit{StaGE} can be seen in Alg.~\ref{alg:method}.

\subsection{Sampling Physically Stable States}\label{sec:samplepts}
In the first stage of our algorithm, we sample a set of fixed stable states $\mathcal{C}_s\subset\mathcal{C}_{stable}$ to guide the search later on. 
To that end, we follow the constrained sampling method introduced in~\cite{toussaint2026constrainedsamplingguideuniversal}, which generates states by solving a non-linear program. We briefly explain the method here and refer the reader to~\cite{toussaint2026constrainedsamplingguideuniversal} for more details.

The method first samples contact variables $c_{ij}\in\{0,1\}$, indicating whether the $i$-th and $j$-th frames in our scene are in contact. For scenes with one underactuated object (scenes (a), (b), and (d) in Fig.~\ref{fig:environments}), we sample $1$ to $3$ support frames. 
If the scene contains multiple objects (scene (c) in Fig.~\ref{fig:environments}), we sample uniformly from the set of up to three contacts, where each contact includes at least one underactuated object. 
Then, for each active contact, we add a point of attack $p_{ij}$, constrained to lie on both surfaces, as well as a force $f_{ij}$ acting on the frames, constrained to be within a predefined friction cone~\cite{toussaint2020describing}. 
Each underactuated object is constrained to be in a quasi-static equilibrium, i.e.~the sum of all forces and moments acting on it through the contacts and gravity is zero.
Lastly, we constrain the state to be collision-free.

We then try to find a state $x$ satisfying these constraints by uniformly sampling a random initial state $\bar x$ formulating the optimization problem 
\begin{equation*}
\begin{aligned}
    \min_{x}\,& \lVert x-\bar x\rVert^2\\
    \text{\textit{subject to}}\,&\left\{\begin{aligned}
        \text{collision}\\
        \text{contact}\\
        \text{force}
    \end{aligned}
    \right\}\!\text{--\,constraints,}
\end{aligned}
\end{equation*}
which we solve using an augmented Lagrangian method~\cite{wright1999numerical}.
This formulation can be understood as a random projection onto the manifold $\mathcal{C}_{stable}$. Some examples of the resulting states can be seen in Fig.~\ref{fig:visualAbstract}.

While this method works well for the simple contact dynamics in the first two scenes in Fig.~\ref{fig:environments}, it can struggle with the more complex meshes of the robotic arm introduced in scenes (c) and (d). That is to say, the optimizer may not find a solution that satisfies the constraint. If so, we sample a random initial state $\bar x$ again and try to solve it until success or a given number of trials is reached. In practice, 
in the \textit{PandaHook} scene, we achieve a $100\%$ success rate in finding a stable state within $100$ attempts when both objects remain in contact with the table. However, imposing a grasp constraint (i.e., enforcing contact between an object and both finger frames of the gripper) reduces performance. The success rate drops to $15.1\%$ when attempting to grasp the cube and to only $1.4\%$ for grasping the hook.
%
To remedy that, we first sort the contacts into high-level abstractions (namely contact with table, with panda or grasps) and then filter our stable states to have roughly equal numbers for each class.
We note that another way to circumvent this problem would be to handle these interactions by using specialized (grasp) samplers~\cite{fang2023anygrasprobustefficientgrasp, sundermeyer2021contactgraspnetefficient6dofgrasp}. However, this is beyond the scope of this work.




\subsection{Connecting States}\label{sec:method-connect-states}

To achieve our goal of finding diverse interactions, we build upon kinodynamic RRT~\cite{lavalle1999randomized}. In its most  basic version, kinodynamic RRT grows a tree rooted at a starting state by uniformly sampling a state and extending the closest node of the tree towards it. The extension works by selecting an action and simulating its effect for a fixed time-step. The end-state of that roll-out is then added as a new leaf of the tree. One possible scheme to select the action is to evaluate random actions through a physical engine and take the action that results in a state minimizing the distance to the sampled target state, as proposed in~\cite{Haustein}. 
In this work, we refer to this approach as RRT-sim.

While being generic,  this approach is insufficient for our use-case. Intuitively, sampling directly in the configuration space will lead to too many uninteresting or irrelevant configurations (e.g., the tool flying in the air).
To circumvent this issue, we propose \textit{stability-guidance}, which means that instead of uniformly sampling from the space $\mathcal{C}_{feasible}$, we only sample from the embedded manifold $\mathcal{C}_{stable}$. Sampling from this manifold is expensive due to the complex, contact-rich interactions. Therefore, we first generate a set of fixed stable states $\mathcal{C}_s$, as described in \cref{sec:samplepts}, from which we draw a point to extend the tree towards. To compute the state to extend from and the action selected for extension, we use the weighted mean squared error as a metric, weighing the position error of non-actuated objects higher than the joint-state error of the robot.

Furthermore, our goal is to find many \textit{diverse} paths between multiple nodes in a dynamic, underactuated system with complex non-prehensile interactions. To improve exploration, we propose three additional extensions:
\begin{itemize}
\item[(i)] \textbf{Sampling from the $K$-Nearest Neighbors}\quad
Instead of taking the node closest to the sampled stable state, we uniformly choose one of its $k$-nearest neighbors. This enables us to grow the tree (and consequently find more paths), even if the closest node is already within the minimum distance to the target state. 
To efficiently compute the $k$-nearest nodes for each stable state, we exploit the fact that the stable states are fixed: each stable state keeps a record of its  $k$-nearest nodes, which we update each time we add a new node to our tree. 
This allows us to execute all the k-nearest neighbor searches and updates during the $N_{\max}$ iterations of the algorithm  with $m=\lvert\mathcal{C}_s\rvert$ stable states in $\mathcal{O}(N_{max}m\log k)$ instead of $\mathcal{O}(N_{max}(\log N_{max} + k))$ when using approximate nearest neighbors.\footnote{in our implementation, the selection step is constant, while the update step costs $\mathcal{O}(m\log k)$ when using a max-heap to save the nearest nodes for each iteration. Using approximate nearest neighbors~\cite{harwood2025approximate, Malkov2020efficient} one could achieve an expected runtime of $\mathcal{O}(\log i + k)$ in the $i$-th iteration, leading to the overall runtime $\mathcal{O}(N_{max}(\log N_{max} + k))$.} Since $N_{max}\gg m$, we use our (exact) implementation. 

\item[(ii)] \textbf{$N$-Best Actions}\quad Selecting the $n$-best actions that reduce the distance to the sampled target state, instead of only choosing the single best one, results in greater diversity in the paths found by our method. 

\item[(iii)] \textbf{Node Rejection}\quad 
If a node fails to expand the tree towards any of the target stable states, we consider it to have a high likelihood of being a dead-end and therefore do not try to expand it further. Our environments contain unrecoverable states. For example, if the sphere falls off the ramp in scene (a) from Fig~\ref{fig:environments} or if the cube is pushed out of reach in (d). We believe that dead-ends act as a task-agnostic proxy for such states.

\end{itemize}


\begin{algorithm}[t]
\caption{StaGE}
\label{alg:method}

\KwIn{Maximum number of nodes $N_{\max}$, number of stable configurations $m$, nearest neighbors $k$, number of best actions $n$, minimal path difference $d_{min}$}
\KwOut{Set of feasible paths $\mathcal{P}$}

\BlankLine
$\mathcal{C}_s \leftarrow \textsc{SampleStableStates}(m)$\;
$x_0 \sim \mathcal{C}_s$\;
\tcp{--- Tree Construction ---}
$\mathcal{T} \leftarrow \textsc{InitializeTree}(x_0)$\;

\For{$i \gets 1$ \KwTo $N_{\max}$}{
    
    $x_{\text{target}} \sim \mathcal{C}_s$\;
    
    $\mathcal{X}_{\text{near}} \leftarrow 
    \textsc{KNearest}(\mathcal{T}, x_{\text{target}}, k)$\;
    $x_{\text{near}} \sim \mathcal{X}_{\text{near}}$\;
    $(\mathcal{A}_n, \mathcal{X}_n) \leftarrow 
    \textsc{OptimizeActions}(x_{\text{near}}, x_{\text{target}}, n)$\;
    \eIf{\textsc{ReducesDistance}$(\mathcal{X}_n, \mathcal{C}_s)$}{
        $\mathcal{T}.\textsc{Expand}(x_{\text{near}}, \mathcal{A}_n, \mathcal{X}_n)$\;
    }{
        $\mathcal{T}.\textsc{Disable}(x_{\text{near}})$\;
    }
}

\BlankLine
\tcp{--- Path Extraction ---}
$\mathcal{P} \leftarrow \emptyset$\;
\ForEach{$x \in \mathcal{T}$}{
    \ForEach{$x_s \in \mathcal{C}_s \setminus \{x_0\}$}{
        \If{$\textsc{Distance}(x, x_s) < \varepsilon$}{
            $p \leftarrow \textsc{PathToRoot}(x)$\; 
            $\mathcal{P} \leftarrow \mathcal{P} \cup \{p\}$\;
        }
    }
}

\BlankLine
$\mathcal{P} \leftarrow \textsc{RemoveRedundant}(\mathcal{P}, d_{min})$\;
\Return{$\mathcal{P}$}\;

\end{algorithm}

\subsection{Extracting Paths}

We create paths from the tree grown in the second stage by selecting all nodes $x$ that fall within a certain minimum distance $\varepsilon$ to any of the stable states and extracting the path from $x_0$ to $x$. 

Afterwards, we filter out redundant paths for each goal state by iterating through the paths in random order and greedily taking a path if its distance to all previously taken paths is greater than a minimum threshold.
We compute the distance between paths (given as sets $P,Q$) using the (undirected) Hausdorff-distance\begin{equation}\label{eq:hausdorff}
\begin{aligned}
    d_H(P,Q) = \max\{\hat d_H(P,Q), \hat d_H(Q,P)\}&\\
    \text{for}\quad\hat d_H(P,Q) = \max_{p\in P}\min_{q\in Q} d(p,q)&\,.
\end{aligned}
\end{equation}




\section{Experiments}
We evaluate our method across four challenging environments. We provide baselines 
and perform ablations of all major algorithmic contributions.

\subsection{Environments}

We use four distinct environments in our experiments, each using different robot morphologies and highlighting different challenges. The environments are depicted in Fig.~\ref{fig:environments}.

\begin{itemize}
    \item[(i)] \textbf{SpheresRamp}\quad The environment consists of a single robot (blue) with a three-dimensional translational joint, manipulating an object (orange) on a ramp with two walls on the sides; a task similar to~\cite{aydinoglu2024consensus}, but on a sloped surface.
    This environment enforces non-prehensile manipulation, since grasps are not possible. Furthermore, it contains non-recoverable states when the ball falls off the ramp. 
    \item[(ii)] \textbf{SpheresCube}\quad This environment contains two robots (blue), each with three-dimensional translational joints, manipulating a cube. The environment contains two walls. This allows for diverse manipulations, where the cube can be pushed, grasped, thrown, or pivoted using the walls and floor. Additionally, the environment also forces the method to find manipulations that change the orientation of the cube.
    \item[(iii)] \textbf{PandaHook}\quad The environment contains a single Franka Emika Panda robotic arm, as provided by~\cite{menagerie2022github}, together with a cuboid and a hook. This environment enables more complex interaction where, for example, the robot can use the hook to reach and manipulate the cube. This environment motivates tool use and demonstrates sequential manipulation planning, where long horizon multi-step manipulations are needed.
    \item[(iv)] \textbf{PandasCube}\quad The environment contains two panda robot arms in a bi-manual setup with a cuboid. This environment encourages collaboration, like throwing the cuboid from one arm to the other.
\end{itemize}

\subsection{Evaluation Metrics}
We evaluate our baselines and method on four different metrics intended to capture the quantity and diversity of the resulting data.
\begin{enumerate}
    \item[(i)] \textbf{Path Count}\quad The total number of diverse paths returned by our method.
    \item[(ii)] \textbf{Coverage}\quad The percentage of stable states sampled at the initial stage, which were reached by at least one node.
    \item[(iii)] \textbf{Entropy}\quad The entropy of the states visited by the paths, calculated using the Kozachenko–Leonenko estimator~\cite{kozachenko1987sample}. This estimator approximates the entropy based on the distance of the $k$-th nearest neighbor.
    We sample $100$ states\footnote{We use a fixed number of states, since the KL-estimator depends on it and the number of visited states varies by multiple orders of magnitude between the methods we compare.} from all visited states and calculate the entropy using $k=10$. We repeat this process $10$ times and report the average.
    \item[(iv)] \textbf{Average Hausdorff}\quad 
    The average pairwise (undirected) Hausdorff distance \eqref{eq:hausdorff} between paths with the same stable state as the endpoint, averaged over all stable end states with at least two paths connecting to it.
\end{enumerate}
For evaluation, we fix the set of stable states for each environment. We evaluate all baselines and ablations using $10$ randomly drawn starting states and average the results. We only report the entropy if the paths returned by the method add up to at least $100$ nodes, and we only report the average Hausdorff-distance if a stable state is reached by at least two paths.

\subsection{Baselines}
We compare our method against kinodynamic RRT through black-box simulation (RRT-sim) introduced in sec.~\ref{sec:method-connect-states}. 
We modify it by adding a bias towards sampling stable states with 20\% probability, which corresponds to \textit{goal-bias} in kinodynamic RRTs~\cite{lavalle1999randomized}.

Additionally, we evaluate the impact of our proposed extensions with ablations. For these, we take the number of selected actions $n\in\{1,16\}$ and the number of nearest neighbors $k\in\{1,16\}$. We also perform an ablation with only a 20\% chance of sampling stable states, the rest of the samples being uniform. We give a budget of 2,500 tree expansions for the \textit{SpheresRamp} and \textit{SpheresCube} environments and a budget of 10,000 tree expansions for the \textit{PandaHook} and \textit{PandasCube} environments. 


Furthermore, we also consider as a baseline sampling-based MPC with predictive sampling. We rely on the implementation provided in Hydrax~\cite{kurtz2024hydrax} on the first (simplest) environment.

\begin{table}[t]
\vspace{0.5cm}
  \centering
  \caption{Experimental results across four metrics.}
  \label{tab:ablations}
  \begin{tabular}{lcccc}
    \toprule
    Method 
      & Count & Cov. & Entropy & Avg. Haus. \\
      &   & [\%] & [nats]  & [$\times10^{-2}$] \\    
    \midrule
    \rowcolor{lightcyan} \multicolumn{5}{c}{\textbf{SpheresRamp}} 
    
    \\
    Predictive Sampling        & $14.3$      & $57.2$      & ---        &  ---
    \\
    RRT-sim                    & $2.5$ & $10$ & --- & --- \\
    StaGE-uniform-80\%         & $30.4$ & $61.6$ & $-27.21$ & $11.27$ \\
    StaGE w/o node rejection   & $50$ & $76.8$ & $-28.38$ & $14.64$ \\
    StaGE w/o n-best actions   & $14.8$ & $37.2$ & $-39.11$ & $17.27$ \\
    StaGE w/o k-NN             & $47$ & $61.2$ & $-30.96$ & $\mathbf{18.13}$ \\
    StaGE                      & $\mathbf{68.8}$ & $\mathbf{85.2}$ & $\mathbf{-26.84}$ & $12.52$ \\
        
    \midrule
    \rowcolor{lightcyan} \multicolumn{5}{c}{\textbf{SpheresCube}} \\
    RRT-sim                    & $0.1$ & $0.1$ & --- & --- \\
    StaGE-uniform-80\%         & $16.1$ & $2.83$ & $-18.66$ & $14.05$ \\
    StaGE w/o node rejection   & $110.8$ & $12.42$ & $\mathbf{-3.37}$ & $14.95$ \\
    StaGE w/o n-best actions   & $0.3$ & $0.2$ & --- & $\mathbf{24.72}$ \\
    StaGE w/o k-NN             & $39.3$ & $7.07$ & $-6.91$ & $19$ \\
    StaGE                      & $\mathbf{134.2}$ & $\mathbf{16.97}$ & $-13.83$ & $15.25$ \\
    
    \midrule
    \rowcolor{lightcyan} \multicolumn{5}{c}{\textbf{PandaHook}} \\
    RRT-sim                    & $0$ & $0$ & --- & --- \\
    StaGE-uniform-80\%         & $4.8$ & $0.36$ & $18.67$ & $31.72$ \\
    StaGE w/o node rejection   & $40.9$ & $1.01$ & $30.3$ & $39.6$ \\
    StaGE w/o n-best actions   & $0$ & $0$ & --- & --- \\
    StaGE w/o k-NN             & $45.4$ & $0.65$ & $31$ & $\mathbf{50.25}$ \\
    StaGE                      & $\mathbf{48.7}$ & $\mathbf{1.44}$ & $\mathbf{37.48}$ & $40.22$ \\

    \midrule
    \rowcolor{lightcyan} \multicolumn{5}{c}{\textbf{PandasCube}} \\
    RRT-sim                    & $3$ & $0.87$ & --- & $54.22$ \\
    StaGE-uniform-80\%         & $24.7$ & $4.45$ & $32.13$ & $55.35$ \\
    StaGE w/o node rejection   & $\mathbf{198.8}$ & $\mathbf{24.68}$ & $51.27$ & $58.86$ \\
    StaGE w/o n-best actions   & $7.4$ & $1.34$ & --- & $53.7$ \\
    StaGE w/o k-NN             & $40$ & $6.02$ & $35.21$ & $\mathbf{66.14}$ \\
    StaGE                      & $170$ & $21.34$ & $\mathbf{52.12}$ & $58.4$ \\
    \bottomrule
  \end{tabular}
\end{table}

\section{Results}
\begin{figure}[t]
  \centering
    \subfloat[RRT-sim]{\includegraphics[width=0.24\textwidth]{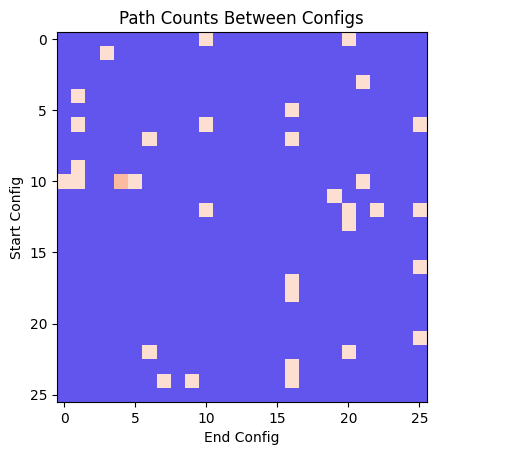}}
    \subfloat[StaGE]{\includegraphics[width=0.24\textwidth]{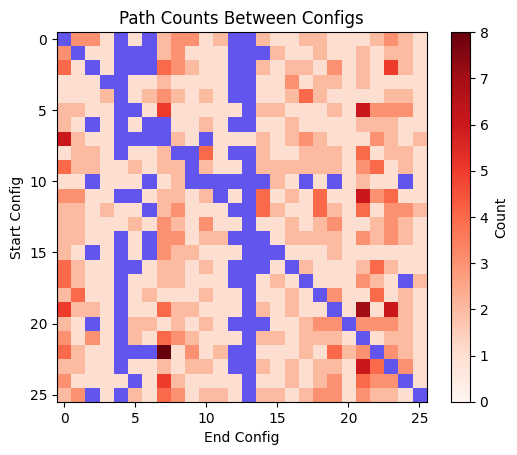}}
  \caption{Adjacency matrices for trajectories found between a set of 26 stable states for the \textit{SpheresRamp} environment. Blue indicates no connections were found between two states, the count value on the right refers to the amount of diverse paths found for each pair of stable states.}
  \label{fig:resk}
\end{figure}

Table~\ref{tab:ablations} shows the results of our experiments. Our method performs the best in terms of coverage and paths found for all environments except \textit{PandasCube}. We hypothesize that the improvement of \textit{StaGE w/o node rejection} over \textit{StaGE} is due to the high dimension of the action space (two $8$-DOF robotic arms). Since many actions (e.g.~null-space movements or movements of the arm not in contact with the box) will not improve the state cost, it is reasonable to assume that the size of the set of actions improving our state is small. Therefore, even on a previously unsuccessful node, actions improving the distance can be sampled in subsequent iterations. 
Future work could investigate alternative sampling schemes for sampling actions, in order to improve their the chances of success.
Fig.~\ref{fig:resk} more clearly shows the ability of our method to explore the space of feasible actions in the \textit{SpheresRamp} environment compared to the \textit{RRT-sim} baseline. Fig.~\ref{fig:trajectories} shows some examples of the trajectories found by \textit{StaGE}.

\begin{figure}[htbp]
    \vspace{0.5cm}
    \centering
    \subfloat[Impact of $n$-best actions on coverage]{\includegraphics[width=\linewidth]{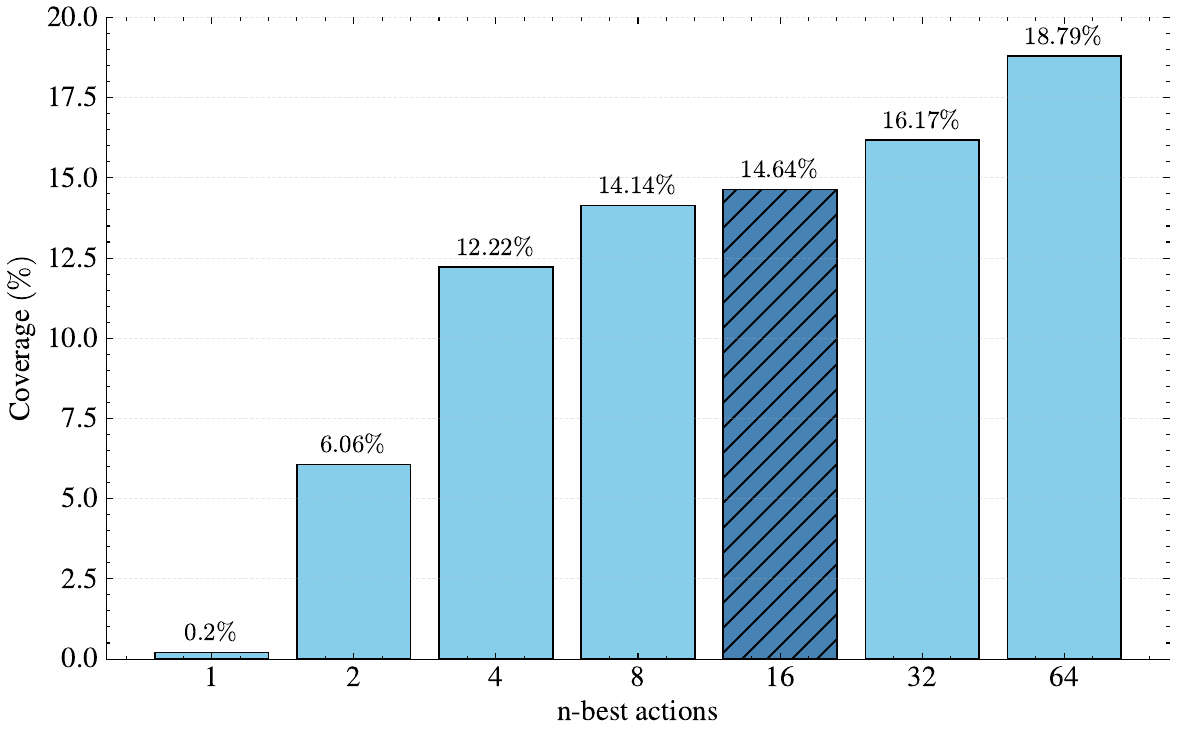}}
    
    \vspace{0.5cm} 
    
    \centering
    \subfloat[Impact of $k$-nearest neighbors on coverage]{\includegraphics[width=\linewidth]{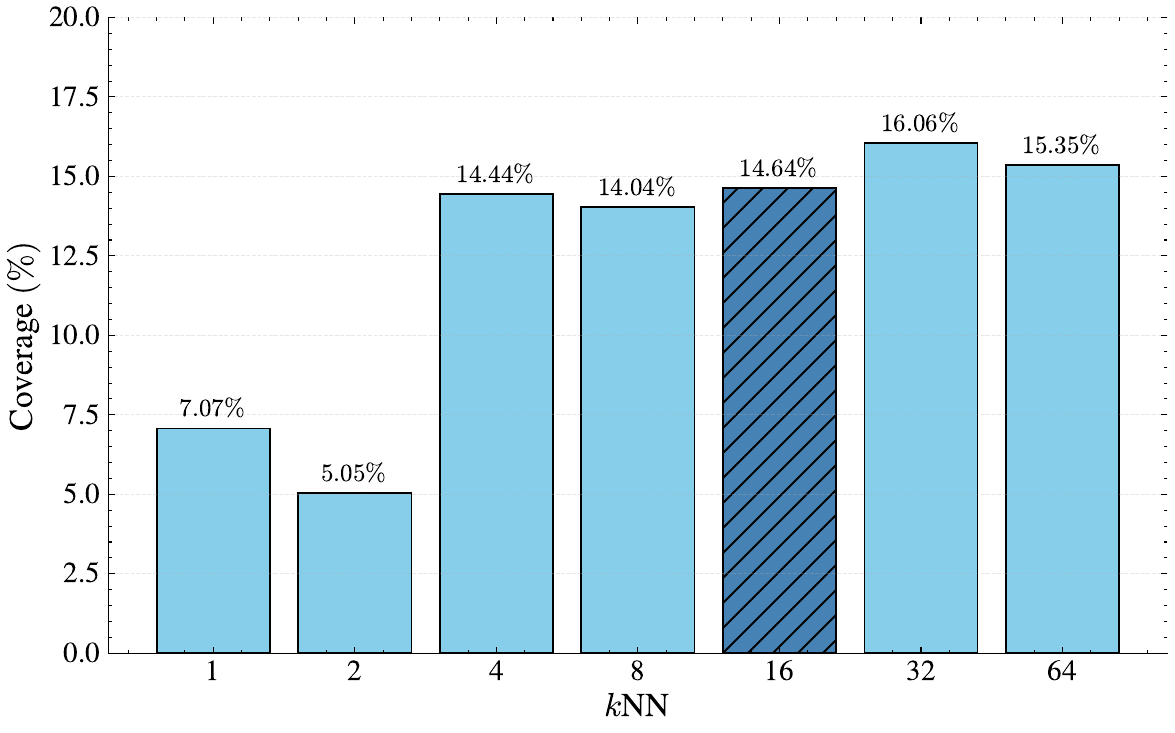}}

    \caption{Ablation study comparing different values for the $n$-best actions and $k$-nearest neighbors parameter on the \textit{SpheresCube} environment for a fixed budged of $2,500$ expansions. We fixed the values for both parameters to $16$ in the rest of our experiments.}
    \label{fig:total_ablation}
\end{figure}

The results demonstrate that taking the $n$-best actions instead of only the best-performing one leads to the biggest improvement in performance. To further investigate the effects of this parameter and the $k$-nearest neighbor parameter we performed the ablations shown in Fig.~\ref{fig:total_ablation}.
Additionally, we performed an ablation on the effects of the number of stable states sampled in the first stage of our method as shown in Table~\ref{tab:stable_counts}. Notice that the coverage metric is highly dependent on the amount of stable states.

Generating the trajectories on an AMD Ryzen 9 9900X CPU takes 4 seconds per trajectory for the \textit{SpheresRamp} scenario, 1 minute per trajectory for \textit{SpheresCube}, 8 minutes per trajectory on \textit{PandaHook}, and 30 seconds per trajectory on \textit{PandasCube}. Significant speedups are possible by parallelizing the simulation using a GPU, which is beyond the scope of this work.

\begin{table}[t]
\vspace{0.5cm}
  \centering
  \caption{Effects for different sizes of the set of stable states $\mathcal{C}_{\text{s}}$.}
  \label{tab:stable_counts}
  \begin{tabular}{lcccc}
    \toprule
    $|\mathcal{C}_{\text{s}}|$  
      & Count & Cov. & Entropy & Avg. Haus. \\
      &   & [\%] & [nats]  & [$\times10^{-2}$] \\    
    \midrule
    \rowcolor{lightcyan} \multicolumn{5}{c}{\textbf{SpheresCube}} \\
    25        & $13.4$ & $8.75$ & $-6.71$ & $14.2$ \\
    50        & $49.5$ & $12.45$ & $-3.71$ & $14.71$ \\
    100       & $134.4$ & $\mathbf{15.86}$ & $\mathbf{-1.82}$ & $15.47$ \\
    250       & $132.7$ & $7.95$ & $-2.65$ & $14.66$ \\
    500       & $175.8$ & $5.75$ & $-4$ & $15.12$ \\
    1000      & $289.1$ & $4.64$ & $-3.59$ & $15.97$ \\
    2500      & $473.6$ & $3.59$ & $-5.94$ & $17.42$ \\
    5000      & $\mathbf{721.3}$ & $2.88$ & $-12.92$ & $\mathbf{17.76}$ \\
    \bottomrule
  \end{tabular}
\end{table}

\section{Discussion}

Our method bears resemblance to the PRM paradigm in the sense that we also first sample points (stable configurations), which we then connect using local planners. However, PRMs usually aim to connect nearest neighbors or nodes within a certain distance~\cite{LaValle2006}. In comparison, we aim to find all possible paths since our focus is on diverse data instead of solving motion planning queries efficiently.
Furthermore, for the considered problem, the construction of a PRM is not trivial. Indeed, in the case of manipulation, building a PRM would involve solving a two-point boundary value problem with non-differentiable dynamics. One could rely on gradient-free optimization. However, this can be costly and can fail in more complex environments.

The action sampling step of the kino-dynamic RRT (and consequently, our method) can be seen as an iteration of predictive sampling~\cite{howell2022predictive} to directly connect stable configuration pairs. It would be interesting to use multiple iterations or more sophisticated gradient-free techniques. However, this is beyond the scope of this paper.

In this work, we employed physically stable states to guide the search for diverse manipulations, motivated by the relative simplicity of generating such states. However, future research should explore extending the set of guidance states to include more generally informative states. For instance, states corresponding to moments of impact between the robot and an object could provide richer structural cues for exploration. Furthermore, improving the smoothness of the generated trajectories remains an important direction for future work.

\section{Conclusion}
This paper introduced \textit{StaGE}, a novel method for diverse motion generation in complex non-prehensile manipulation scenarios. We demonstrated that our approach generalizes to a wide range of novel environments without requiring task-specific guidance, showing that pure exploration alone can yield long-horizon sequential manipulation behaviors. Even in the absence of motion priors, the method discovers highly complex skills such as throwing, grasping, pivoting, pushing, handovers, and tool use, relying solely on stable states as guidance.


\section*{Acknowledgments}
This work has received support from the German Federal Ministry of Research, Technology and Space (BMFTR) under the Robotics Institute Germany (RIG).

\bibliographystyle{IEEEtranDOI} 
\bibliography{references}
\end{document}